\documentclass[10pt,conference]{IEEEtran}

\newif\ifieee
\ieeefalse

\usepackage[sort,compress]{cite}
\newcommand{\miniscule}{\fontsize{4}{5}\selectfont}

\ifCLASSINFOpdf
   \usepackage[pdftex]{graphicx}
   \graphicspath{{figs/}}
   \DeclareGraphicsExtensions{.pdf,.jpeg,.png}
\else
   \usepackage[dvips]{graphicx}
   \graphicspath{{../figs/}}
   \DeclareGraphicsExtensions{.eps}
\fi

\usepackage[cmex10]{amsmath}
\usepackage[dvipsnames,table,xcdraw]{xcolor}
\usepackage{amsthm}

\usepackage{lipsum}  
\usepackage{algorithmicx}
\usepackage[ruled]{algorithm}
\usepackage{algpseudocode}

\usepackage{array}

\ifCLASSOPTIONcompsoc
  \usepackage[caption=false,font=normalsize,labelfont=sf,textfont=sf,farskip=0pt]{subfig}
\else
  \usepackage[caption=false,font=footnotesize,farskip=0pt]{subfig}
\fi

\usepackage[hyphens]{url}

\usepackage{xcolor}
\usepackage{amsfonts} 
\usepackage{booktabs}

\newif\iffinal
\finaltrue
\newcommand{\cmtid}{43}
\newcommand{\papertitle}{Multi-Feature Aggregation in Diffusion Models\\for Enhanced Face Super-Resolution} %

\iffinal
\else
\usepackage[switch]{lineno}
\fi

\usepackage[normalem]{ulem}

 \newcommand*{\MS}[2][]{\textcolor{purple}{[\textbf{\ifthenelse{\equal{#1}{}}{MS}{MS(#1)}}: #2]}}

\newcommand*{\DM}[2][]{\textcolor{ForestGreen}{[\textbf{\ifthenelse{\equal{#1}{}}{DM}{DM(#1)}}: #2]}}
\newcommand*{\RL}[2][]{\textcolor{Rhodamine}{[\textbf{\ifthenelse{\equal{#1}{}}{RL}{RL(#1)}}: #2]}}
\newcommand*{\JN}[2][]{\textcolor{blue}{[\textbf{\ifthenelse{\equal{#1}{}}{JN}{JN(#1)}}: #2]}}

\ifieee
\else
\usepackage[colorlinks=true,allcolors=black]{hyperref}
\fi
\usepackage[acronym]{glossaries}
\usepackage{xspace}
\usepackage[capitalise]{cleveref} %
\usepackage[T1]{fontenc}
\usepackage{balance}

\newacronymstyle{long-short-br}
{%
  \GlsUseAcrEntryDispStyle{long-short}%
}%
{%
  \GlsUseAcrStyleDefs{long-short}%
}
\setacronymstyle{long-short-br}

\usepackage{transparent}
\usepackage{tikz}
\newcommand\copyrighttext{%
  \scriptsize Accepted at SIBGRAPI 2024. The final published version is available on IEEE Xplore (DOI: \href{https://doi.org/10.1109/SIBGRAPI62404.2024.10716316}{\textcolor{blue}{10.1109/SIBGRAPI62404.2024.10716316}}).}
\newcommand\copyrightnotice{%
\begin{tikzpicture}[remember picture,overlay]
\node[anchor=south,yshift=30pt,xshift=0pt] at (current page.south) {\fbox{\transparent{0.85}\parbox{\dimexpr0.77\textwidth-\fboxsep-\fboxrule\relax}{\copyrighttext}}};
\end{tikzpicture}%
}

\begin{document}
\title{\huge\papertitle}

\iffinal

\author{
\begin{tabular}{cc}
\multicolumn{2}{c}{
\hspace{1.5mm}Marcelo dos Santos\IEEEauthorrefmark{1}, Rayson Laroca\IEEEauthorrefmark{2}$^,$\IEEEauthorrefmark{1}, Rafael O. Ribeiro\IEEEauthorrefmark{3}, Jo\~{a}o C. Neves\IEEEauthorrefmark{4}, David Menotti\IEEEauthorrefmark{1}} \\[0.5ex] \hspace{8.5mm}\small
\IEEEauthorrefmark{1}\hspace{0.15mm}Federal University of Paran\'a, Curitiba, Brazil & \small \IEEEauthorrefmark{2}\hspace{0.15mm}Pontifical Catholic University of Paran\'a, Curitiba, Brazil \\[-2.5pt]
\hspace{8.5mm}\small \IEEEauthorrefmark{3}\hspace{0.15mm}Brazilian Federal Police, Bras\'{\i}lia, Brazil & \small \IEEEauthorrefmark{4}\hspace{0.15mm}University of Beira Interior, Covilh\~{a}, Portugal\ \\
\multicolumn{2}{c}{\resizebox{0.9\linewidth}{!}{
\hspace{1.5mm}\IEEEauthorrefmark{1}{\hspace{-0.35mm}\tt\normalsize \{msantos,menotti\}@inf.ufpr.br} \; \IEEEauthorrefmark{2}{\hspace{0.15mm}\tt\normalsize rayson@ppgia.pucpr.br} \; \IEEEauthorrefmark{3}{\tt\normalsize rafael.ror@pf.gov.br}  \; \IEEEauthorrefmark{4}{\tt\normalsize jcneves@di.ubi.pt}
}}
\end{tabular}
}

\else
  \author{SIBGRAPI paper ID: \cmtid \\[10ex] }
  \linenumbers
\fi

\maketitle

\ifieee
    {\let\thefootnote\relax\footnote{\\979-8-3503-7603-6/24/\$31.00
    \textcopyright2024 IEEE}}
\else
    \copyrightnotice
\fi

\newacronym{ddpm}{DDPM}{Denoising Diffusion Probabilistic Models}
\newacronym{ffhq}{FFHQ}{Flickr-Faces-HQ}
\newacronym{gan}{GAN}{Generative Adversarial Network}
\newacronym{hr}{HR}{high-resolution}
\newacronym{mcmc}{MCMC}{Markov chain Monte Carlo}
\newacronym{mse}{MSE}{Mean Squared Error}
\newacronym{psnr}{PSNR}{peak signal-to-noise ratio}
\newacronym{smld}{SMLD}{Score matching with Langevin dynamics}
\newacronym{sr}{SR}{super-resolution}
\newacronym{lr}{LR}{low-resolution}
\newacronym{ssim}{SSIM}{structural similarity index measure}
\newacronym{sde}{SDE}{Stochastic Differential Equation}
\newacronym{sdes}{SDEs}{Stochastic Differential Equations}
\newacronym{ve}{VE}{variation exploding}
\newacronym{vp}{VP}{variation preserving}
\newacronym{srname}{FASR}{Feature Aggregation Super-Resolution}
\newcommand{\ffhq}{\gls*{ffhq}\xspace}
\newcommand{\gfpgan}{GFP-GAN\xspace}
\newcommand{\sparnet}{SPARNet\xspace}

\iffinal
\newcommand{\supplementary}{\url{https://github.com/marcelowds/fasr}}

\else

\newcommand{\supplementary}{[hidden for review]}

\fi
\ifieee
\vspace{-3.575mm}
\else
\vspace{-3.575mm}
\fi
\begin{abstract}
Super-resolution algorithms often struggle with images from surveillance environments due to adverse conditions such as unknown degradation, variations in pose, irregular illumination, and occlusions.
However, acquiring multiple images, even of low quality, is possible with surveillance cameras.
In this work, we develop an algorithm based on diffusion models that utilize a low-resolution image combined with features extracted from multiple low-quality images to generate a super-resolved image while minimizing distortions in the individual's identity.

Unlike other algorithms, our approach recovers facial features without explicitly providing attribute information or without the need to calculate a gradient of a function during the reconstruction process.
To the best of our knowledge, this is the first time multi-features combined with low-resolution images are used as conditioners to generate more reliable super-resolution images using stochastic differential equations. The FFHQ dataset was employed for training, resulting in state-of-the-art performance in facial recognition and verification metrics when evaluated on the CelebA and Quis-Campi datasets.
Our code is publicly available at \textit{\supplementary}.
\end{abstract}

\IEEEpeerreviewmaketitle

\section{Introduction}
\label{sec:introduction}

\glsresetall

 \begin{figure*}[!ht]
\centering
\hspace{-4mm}\includegraphics[width=0.98\linewidth]{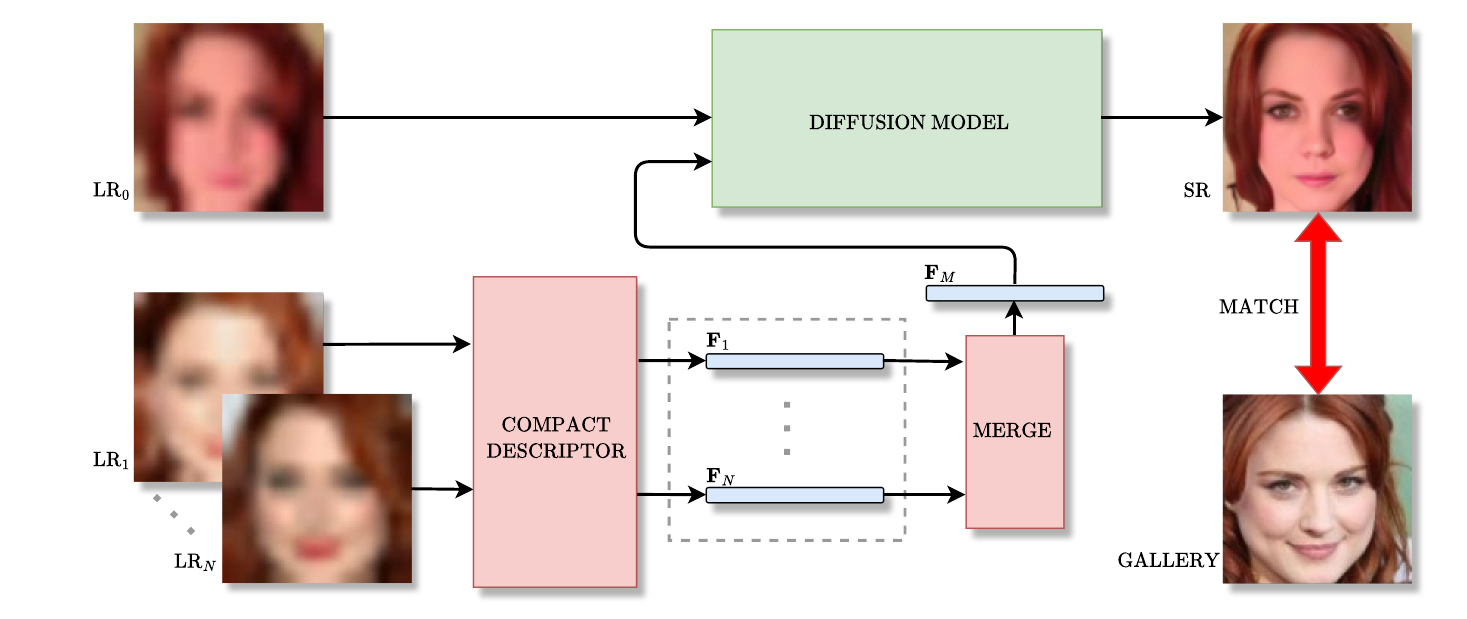}
\vspace{-0mm}
\caption{
Overview of the proposed method. The low-resolution images $\text{LR}_1, \dots, \text{LR}_N$ are used to compute a set of features $\mathbf{F}_1,\dots,\mathbf{F}_N$, respectively, which are then combined to generate $\mathbf{F}_M$. The low-resolution image $\text{LR}_0$ is integrated with $\mathbf{F}_M$ in the diffusion model to produce a super-resolution (SR) image. The SR image is subsequently compared with a set of images from the gallery for face recognition.
}
\label{fig_overview}
\end{figure*}

The problem of \gls*{sr} is inherently ill-posed, making the recovery of fine details like eyeglasses, beards, mustaches, and a reliable identity quite challenging \cite{ill-2002, jiang2021deep}.
For surveillance scenarios, the presence of noise, occlusions, variations in illumination, and varying poses make the problem even harder, leading to a significant decline in the performance of \gls*{sr} and face recognition algorithms \cite{chag-surv-2016,nascimento2022combining}.

Soft biometrics, such as gender, hair color, and skin tone, can enhance image reconstruction, reducing the ambiguity in face \gls*{sr} and increasing the reliability of recognition systems~\cite{yu2018super}.
However, facial attributes are often not visible in \gls*{lr} images, making reliable access challenging.
Also, obtaining these attributes typically requires a classifier or manual extraction, which is not very efficient~\cite{visapp24sr}.

In \cite{yu2018super} and \cite{visapp24sr}, attributes such as beard and glasses, among others, are used to improve the quality of \gls*{sr} algorithms.
Nevertheless, these attributes alone are insufficient to generate accurate high-resolution images.
It is also necessary to consider subtle characteristics, such as facial proportions, shapes, and other high-level, more abstract features.
Therefore, it is essential to develop algorithms that rely on more general characteristics as sources of information for image~reconstruction.

Diffusion models are used for data generation across diverse domains, and here, they are employed to generate \gls*{sr} images.
These models operate by adding noise at different scales to the data and training a network to predict the noise present.
Once trained, the network can perform reverse diffusion, removing the noise and generating the desired type of~data.

Diffusion models can integrate various types of information, such as text and image~\cite{saharia2022photorealistic}, image and audio~\cite{richter2023audio}, and multi-modal data~\cite{huang2023collaborative}.
This enables the generation of extraordinarily high-quality and original data. 
This concept is central to our work, where we combine \gls*{lr} images with facial~features.

Another tool commonly used in conjunction with diffusion models is the classifier guidance method, which is used to generate data within predefined classes or with specific characteristics.
It involves utilizing a classifier's gradient as a supervisor during reverse diffusion. 
For instance, it was used in~\cite{visapp24sr} to recover facial attributes.
A drawback of this strategy is the need to train a classifier, along with the additional computational cost associated with gradient~calculations.

The main contribution of our work lies in developing \gls*{srname}, an \gls*{sr} algorithm that recovers crucial features for face recognition.
In addition to the \gls*{lr} image, it takes as input a vector of facial features derived from a set of \gls*{lr} images, which can be either a series of video frames or independent images of an individual. This new vector has a higher signal-to-noise ratio than each vector individually. It is incorporated into the network, merging its information with \gls*{lr} image to generate an \gls*{sr} version.
In this way, our algorithm effectively recovers facial information from an image, yielding results of higher quality with minimal distortion of identity.
\gls*{srname} employs diffusion models based on~a~\gls*{sde} and operates without the need for a classifier to guide the reverse diffusion~process.

Our method's effectiveness has been validated on the CelebA and QuisCampi datasets. Our SR algorithm produces superior qualitative and quantitative results.
The state-of-the-art values are supported by better results in face recognition metrics such as Area Under the Curve (AUC) in the 1:1 verification protocol and accuracy in the 1:N identification~protocol.

This paper is structured as follows.
Section~\ref{sec:related} outlines related works, Section~\ref{sec:Theoretical} describes the main concept behind the proposed method, and Section~\ref{sec:Exp_and_Res} details our experiments and results.
Finally, Section~\ref{sec:Conclusions} concludes the~paper.

\section{Related Work}
\label{sec:related}

In the seminal work \cite{sohl2015deep},  Sohl-Dickstein et al. utilized principles from non-equilibrium thermodynamics to create a generative model. Two other works in the line of diffusion models that had much impact in this field are Denoising Diffusion Probabilistic Models (DDPMs) \cite{ho2020denoising} and Score-Based Generative Models (SGMs) \cite{song2019generative, song2020improved}.
In \cite{song2021score}, DDPM and SGD are generalized for continuous time steps and noise levels using \gls*{sdes}, expanding the range of possibilities of research in diffusion~models.

Due to the rapid evolution of diffusion models, various opportunities for their application have emerged. Recent works include the generation of audio, graphs, and shapes, as well as image synthesis, solutions of general inverse problems, and applications in medical images \cite{niu2020permutation, cai2020learning, ho2020denoising, song2019generative, song2021score, song2022solving}.
The full potential of diffusion models can also be leveraged through multi-domain data integration, such as text-to-image translation~\cite{saharia2022photorealistic} and image editing~\cite{zhang2023sine}.
Additionally, \cite{richter2023audio} combines audio-visual information for speech enhancement.

\gls*{sr} is another significant application of diffusion models, which is utilized in this work.
In~\cite{saharia2021image}, an adaptation of the DDPM model produces high-quality \gls*{sr} images.
Similarly, SRDiff~\cite{li2022srdiff} employs diffusion models to estimate the difference between the original \gls*{lr} image and an \gls*{hr} image, resulting in an \gls*{sr} image.
In~\cite{dos2022face}, \glspl*{sde} were used to generate \gls*{sr} images.
Additionally, \cite{visapp24sr} performs \gls*{sr} by incorporating attribute information such as beard, gender, and the presence of eyeglasses to generate high-quality images.
However, this approach has the drawback that these attributes must be explicitly provided to the algorithm, which cannot be easily estimated in \gls*{lr}~images.

In \cite{suin2024diffuse}, an identity-preserving \gls*{sr} method was developed.
In both~\cite{visapp24sr} and \cite{suin2024diffuse}, a gradient must be calculated during the image reconstruction phase, which can increase computational cost. In this study, we develop an algorithm that restores image attributes by supplying a compact descriptor of facial features for the~algorithm.

Despite the excellent results achieved by diffusion models, a major drawback is their high execution time due to their iterative nature.
However, this issue is likely to be mitigated in the mid-term, as several studies are focused on enhancing the computational efficiency of these methods.
For a more detailed discussion on accelerating and improving the efficiency of sampling in diffusion models, refer to~\cite{jolicoeur2021gotta, vahdat2021score, meng2023distillation}.

\section{Proposed Method}
\label{sec:Theoretical}

In this section, we present the general idea of our proposed method, followed by a brief theoretical background on diffusion models based on \gls*{sdes} and a description of how the facial features are incorporated into the model.

\subsection{General Idea}
As previously noted, images captured in surveillance environments are often low-quality.
Nevertheless, in certain instances, a video of a particular person can provide multiple \gls*{lr} images that can be combined to enhance the performance of \gls*{sr} algorithms.
This combination of information from multiple images is expected to increase the signal-to-noise ratio, providing higher-quality information.

In this study, we aim to improve the performance of an \gls*{sr} algorithm by integrating a reference \gls*{lr} image with a combination of multiple features extracted from different \gls*{lr} images (see Fig.~\ref{fig_overview}).
This integration leads to enhanced image reliability concerning person identification.
Moreover, the algorithm successfully retrieves high-level features that might not be clearly visible but significantly enhance recognition accuracy and image~quality.

\subsection{Theoretical Background}

In the context of image generation, diffusion models have two phases: forward diffusion and reverse diffusion.
During forward diffusion, Gaussian noise is added to the image, and a network is trained to predict this noise.
In reverse diffusion, an image composed purely of noise is iteratively denoised and transformed into an image that follows a distribution similar to the images in the training set.
If the diffusion procedure is continuous, it can be modeled using an~\gls*{sde}.

According to \cite{song2021score, anderson1982reverse}, a forward diffusion process $\{\mathbf{x}(t)\}_{t=0}^T$ and its reverse are, respectively, modeled using the following \glspl*{sde}:
\begin{equation}
    \mathrm{d}\mathbf{x}=\mathbf{f}(\mathbf{x},t)\mathrm{d}t+g(t)\mathrm{d}\mathbf{w},
    \label{eqsde}
\end{equation}
\begin{equation}
    \mathrm{d}\mathbf{x}=[\mathbf{f}(\mathbf{x},t)-g(t)^2\nabla_{\mathbf{x}}\log p_t(\mathbf{x})]\mathrm{d}t+g(t)\mathrm{d}\bar{\mathbf{w}},
    \label{revsde}
\end{equation}
where $\mathbf{f}(\mathbf{x},t)$ is the drift coefficient, $g(t)$ is a diffusion coefficient,  $\mathbf{w}$ and $\bar{\mathbf{w}}$ are Wiener process (the former runs backward in time) and $p_t$ is the probability density of $\mathbf{x}(t)$.  References~\cite{sdeplaten, sarkka2019applied} supply more details about Itô \gls*{sde}s and the Wiener~process.

Here, we consider $\mathbf{x}_t$ as an image to be denoised. At $t=0$, the noise level in the image is zero, and at $t=T$, the noise is at its maximum, and there is no information on the image. To obtain an SR image, we need to solve Equation \ref{revsde}, and for that, we use a deep neural network $s_\theta$ to approximate $\nabla_{\mathbf{x}}\log p_t(\mathbf{x})$. The neural network is conditioned on \gls*{lr} images and image features, denoted by $\mathbf{y}$ and $\mathbf{F}_M$, respectively. The training of the neural network  $s_{\theta}(\mathbf{x}(t),\mathbf{y},\mathbf{F}_M,t)$ is achieved by  optimizing the following loss function~\cite{vincent2011connection}:
\begin{align}
     \min_{\theta}
    \mathbb{E}_{t\sim \mathcal{U}[0,T]}
   \mathbb{E}_{\mathbf{x}_0\sim p(\mathbf{x}_0)}\mathbb{E}_{\mathbf{x}(t)\sim p_t(\mathbf{x}(t)|\mathbf{x}(0)}
    \big[\lambda(t)\quad\nonumber\\ \times\, \|s_\theta(\mathbf{x}(t),\mathbf{y},\mathbf{F}_M,t)-\nabla_{\mathbf{x}(t)}\log p_{}(\mathbf{x}(t)|\mathbf{x}(0))\|_2^2\big],
    \label{eqloss}
\end{align}
where $\lambda(t)$ is a positive weighting function and $p_{}(\mathbf{x}(t)|\mathbf{x}(0))$ is the transition kernel from $\mathbf{x}(0)$ to $\mathbf{x}(t)$. 

Here, we use the  \gls*{ve}  case described in~\cite{song2021score} with $\mathbf{f}(\mathbf{x},t)$ and $g(t)$ given respectively by:
\begin{equation}
\mathbf{f}(\mathbf{x},t)=\mathbf{0},\quad g(t)=\sqrt{\frac{\mathrm{d}\sigma^2(t)}{\mathrm{d}t}},    
\end{equation}
\noindent where $\sigma(t)=\sigma_{min}\left(\sigma_{max}/\sigma_{min}\right)^t$ denotes the noise level of the image at the time $t$.

For $\mathbf{f}(\mathbf{x},t)$ and $g(t)$ described above, the mean and variance of $p(\mathbf{x}(t)|\mathbf{x}(0))$ are given by~\cite{song2021score}:
\begin{equation}
\pmb{\mu}(t) = \mathbf{x}(0), \,\, \pmb{\Sigma}(t)=[\sigma^2(t)-\sigma^2(0)]\mathbf{I}.  
\label{eq_mean}
\end{equation}
Thus, we can analytically compute $\nabla_{\mathbf{x}}\log p(\mathbf{x}(t)|\mathbf{x}(0))$ in Equation~\ref{eqloss},  allowing for efficient model training.
Once the network well estimates the gradient, we change  $\nabla_{\mathbf{x}}\log p_t(\mathbf{x})$  by $s_{\theta}(\mathbf{x}(t),t)$ in the reverse process (Equation~\ref{revsde}) and solve it from $t=T$ to $t=0$ using the Euler-Maruyama method~\cite{sdeplaten, sarkka2019applied} to generate an SR image~$\mathbf{x}(0)$.

\subsection{Model Conditioning}

As in most diffusion models, we employ the U-Net architecture~\cite{ho2020denoising}.
To condition the model on LR images, we follow a method similar to the one outlined in~\cite{saharia2021image,dos2022face}.
This method involves concatenating in the channel domain, the LR image~$\mathbf{y}$ and~$\mathbf{x}_T$, which is the image undergoing denoising.
This concatenation results in a 6-channel~image.

\begin{figure}[!htb]
\centering
\includegraphics[width=0.88\linewidth]{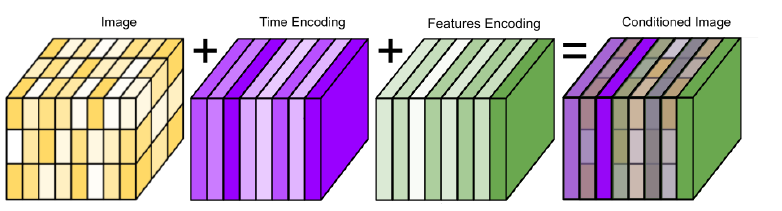}

\vspace{-2mm}
\caption{Time and features encoding. Adapted from \cite{assemblyai_imagen}.}
\label{fig_embed}
\end{figure}

The network is conditioned on image features using a method similar to time and class conditioning described in~\cite{nichol2021improved}.
For a given level of the layers, the weighted sum of the time embedding and the feature vector is added to the image, providing a conditioned image, as shown in Fig.~\ref{fig_embed}.

We used a compact descriptor (see Section~\ref{sec_exp} for more details) extracted from the HR images during training to condition the neural network. The LR images are upsampled to the original size of HR images to preserve the dimensions.
During the evaluation phase, one $\text{LR}_0$ image is used as a reference (in general, one should select the best and most frontal image as the $\text{LR}_0$ reference image, but here it was chosen randomly), while other images $\text{LR}_1,\dots ,\text{LR}_N$ are used to compute feature vectors $\mathbf{F}_1,\dots ,\mathbf{F}_N$. The merging of feature vectors can be performed in various ways. In this work, the merged $\mathbf{F}_M$ is the arithmetic mean of all feature vectors $\mathbf{F}_1,\dots ,\mathbf{F}_N$. We are assuming that $\mathbf{F}_M$ has a higher signal-to-noise ratio than a feature vector obtained from an individual \gls*{lr} image and that this mean vector approximately represents the actual characteristics of the HR image.
\begin{figure}[!ht]
\centering
\setlength{\tabcolsep}{1.2pt}
\resizebox{0.85\linewidth}{!}{ %
\begin{tabular}{cccc}
\includegraphics[width=15mm]{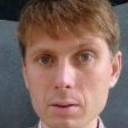} & 
\includegraphics[width=15mm]{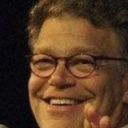} &\includegraphics[width=15mm]{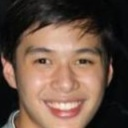}
&\includegraphics[width=15mm]{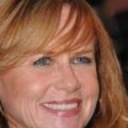}\\[-0.3ex]
\includegraphics[width=15mm]{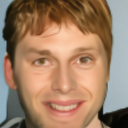} & 
\includegraphics[width=15mm]{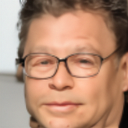}&
\includegraphics[width=15mm]{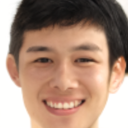}&
\includegraphics[width=15mm]{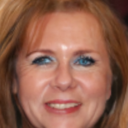}\\

\end{tabular}
}
\vspace{-1mm}
\caption{First row: original HR images extracted from the CelebA dataset. Second row: synthetic HR images generated using only the feature vector extracted from the corresponding image~above.}
\label{fig_from_fv}
\end{figure}

To demonstrate the efficacy of utilizing the feature vector to generate an SR image, we trained Equation~\ref{eqloss} with $\mathbf{y}=0$ and exclusively employed the feature vector to produce some images, as depicted in Fig.~\ref{fig_from_fv}.
These images showcase the algorithm's ability to reconstruct crucial high-level features necessary for preserving a person's identity.
When coupled with the \gls*{lr} image, the feature vector proves effective in restoring image details while minimizing identity distortion.

\newpage
\section{Experiments and Results}
\label{sec:Exp_and_Res}
This section describes the experimental setup, followed by the results obtained on two distinct datasets. Lastly, we examine some extreme cases where the algorithm may fail.
\subsection{Experiments}
\label{sec_exp}

In this study, we explored three datasets: FFHQ~\cite{karras2019style}, CelebA~\cite{liu2015faceattributes}, and Quis-Campi~\cite{neves2018quis}, all gathered from surveillance scenarios.
The FFHQ dataset was employed for model training, where $10^6$ training steps were conducted.
CelebA was employed to test our approach, with $500$ identities selected.
Each identity comprises multiple images, with one randomly chosen as the gallery image.
A second image is downsampled to create a \gls*{lr} probe image.
The remaining images were also downsampled and used to extract features, assisting the reconstruction process of the LR probe~image.

A complementary test to further validate our algorithm was conducted on a real-world scenario from the Quis-Campi dataset, where the images pose additional challenges for SR and face recognition algorithms.
We selected $90$ identities and used five downsampled images as probe images for each identity.
These images were then used to calculate an average feature vector, which was utilized to generate the SR images.
In addition, the dataset already contains gallery images obtained in a controlled environment for each identity. %

The parameters controlling the noise level over time were set at $\sigma_{min}=0.001$ and $\sigma_{max} = 348$.
We worked with images of $128 \times 128$ pixels.
For producing LR images, we applied $8\times8$ downsampling followed by upsampling using bicubic interpolation to achieve a final size of $128 \times 128$ pixels.
We used $2{,}000$ steps to solve the SDE for image~reconstruction.

The feature vector used for training the SR algorithm and for facial recognition consists of a 512-dimensional vector generated through AdaFace~\cite{kim2022adaface} with a ResNet backbone~\cite{he2016deep} trained on the CASIA-WebFace dataset~\cite{yi2014learning}. Image descriptors were compared using the cosine similarity metric. For the recognition task, we compare the SR-recovered images against the gallery images. Our proposed algorithm is compared against state-of-the-art algorithms: SR3~\cite{saharia2021image} and SDE-SR~\cite{dos2022face}.

\subsection{Results}

Table \ref{tab1} shows the quantitative results of our algorithm on the CelebA dataset.
Notably, \gls*{srname} provides superior performance in terms of AUC, Rank-1, Rank-5, and Rank-10 (with an improvement of up to $4\%$) compared to other algorithms.
As our algorithm incorporates the feature vector during the image generation phase, the images can be recovered while maintaining features that positively impact recognition and verification.
\begin{table}[!ht]
\centering
\caption{The 1:1 verification and 1:N identification (Rank-1, Rank-5 and Rank-10) results obtained using the AdaFace recognition model through super-resolution on the CelebA dataset. 
}
\vspace{-2mm}
\begin{tabular}{c c c c c}
\toprule
 SR Method & \multicolumn{1}{c}{AUC} &   \multicolumn{1}{c}{Rank-$1$ (\%)}&  \multicolumn{1}{c}{Rank-$5$ (\%)}&  \multicolumn{1}{c}{Rank-$10$ (\%)}\\ \midrule
 LR &  $0.885$  &  $27.00$   &$41.40$   &$51.60$  \\
  SR3 &  $0.936$  &  $45.60$   &$62.00$   &$71.00$  \\
 SDE-SR & \cellcolor{gray!15}  $\!\!0.933$  &\cellcolor{gray!15}${48.60}$  &\cellcolor{gray!15}${66.60}$  & \cellcolor{gray!15}  $\!\!72.40$  \\
\gls*{srname}~(Ours) & \cellcolor{gray!30}${0.946}$  &\cellcolor{gray!30}${52.80}$  &\cellcolor{gray!30}${70.00}$  &\cellcolor{gray!30}${76.00}$\\
\bottomrule
\end{tabular}
\label{tab1}
\vspace{-1mm}
\end{table}
Table~\ref{tab2} shows the quantitative results on the Quis-Campi dataset.
In this context, we made additional comparisons with IDM~\cite{gao2023implicit} and SRGD~\cite{visapp24sr}, state-of-the-art algorithms.
IDM represents an enhancement of SR3, while SRDG takes feature information as input and attempts to incorporate these features in SR images.
Our algorithm provides superior results in recognition accuracy~(Rank-$1$).
\begin{table}[!ht]
\centering
\caption{The 1:1 verification and 1:N identification (Rank-1, Rank-5 and Rank-10) results obtained using the AdaFace recognition model through super-resolution on the Quis-Campi~dataset.
}
\vspace{-2mm}
\begin{tabular}{c c c c c}
\toprule
 SR Method & \multicolumn{1}{c}{AUC} &   \multicolumn{1}{c}{Rank-1(\%)}&  \multicolumn{1}{c}{Rank-5(\%)}&  \multicolumn{1}{c}{Rank-10(\%)}\\ 
 \midrule
 LR & $0.816$  &  $23.78$  &  $46.89$  &  $58.67$  \\
  IDM & $0.885$  &  $28.22$  &   $56.44$  &   $70.00$  \\
 SR3 & $0.914$  & $45.78$  &   $69.56$  &   $79.77$  \\
 SDE-SR &\cellcolor{gray!0}  $\!\!0.917$  &  \cellcolor{gray!15}${50.00}$  & \cellcolor{gray!15}${72.67}$  &  \cellcolor{gray!15}  $\!\!81.56$  \\ 
SRDG  & \cellcolor{gray!30}${0.920}$  &  \cellcolor{gray!0}${49.33}$  &  \cellcolor{gray!30}${73.11}$  &  \cellcolor{gray!30}${82.00}$ \\
\gls*{srname} (Ours) & \cellcolor{gray!15}${0.917}$  &  \cellcolor{gray!30}${51.33}$  &  \cellcolor{gray!0}${72.44}$  &  
\cellcolor{gray!0}${80.00}$  \\
\bottomrule
\end{tabular}
\label{tab2}
\vspace{-1mm}
\end{table}

\begin{figure*}[!ht]
\centering
\setlength{\tabcolsep}{1pt}
\resizebox{\linewidth}{!}{ %
\begin{tabular}{ccccccc}
\miniscule{LR}&\miniscule{SR3~\cite{saharia2021image}}&\miniscule{IDM~\cite{gao2023implicit}}&\miniscule{SDE-SR~\cite{dos2022face}}&\miniscule{SRDG~\cite{visapp24sr}}&\miniscule{\gls*{srname} (Ours)}&\miniscule{GT}\\[-0.25ex]
 \includegraphics[width=10mm]{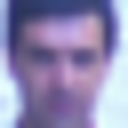} &  \includegraphics[width=10mm]{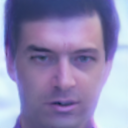}&  \includegraphics[width=10mm]{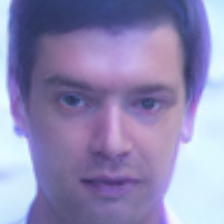}&\includegraphics[width=10mm]{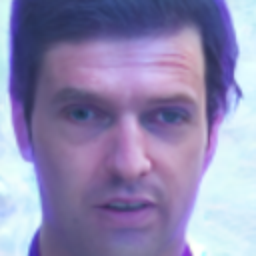}   &  \includegraphics[width=10mm]{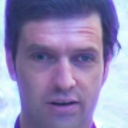} &  \includegraphics[width=10mm]{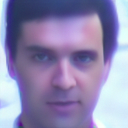}&\includegraphics[width=10mm]{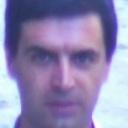}  \\[-0.4ex]
 \includegraphics[width=10mm]{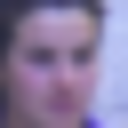} &  \includegraphics[width=10mm]{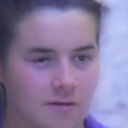}&  \includegraphics[width=10mm]{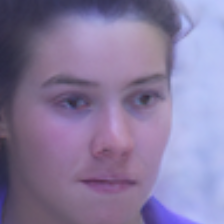}&\includegraphics[width=10mm]{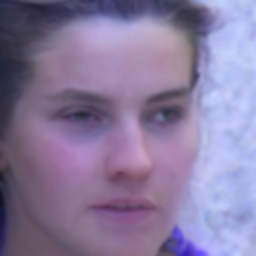}   &  \includegraphics[width=10mm]{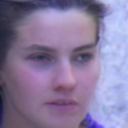} &  \includegraphics[width=10mm]{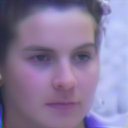} &\includegraphics[width=10mm]{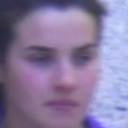} \\[-0.4ex]
 \includegraphics[width=10mm]{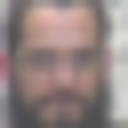} &  \includegraphics[width=10mm]{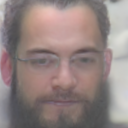}&  \includegraphics[width=10mm]{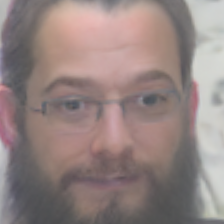}&\includegraphics[width=10mm]{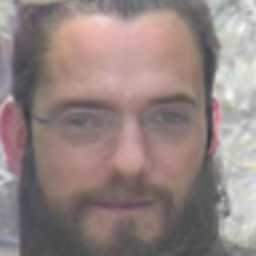}   &  \includegraphics[width=10mm]{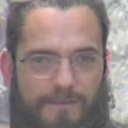} &  \includegraphics[width=10mm]{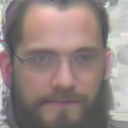}&\includegraphics[width=10mm]{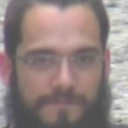}  \\[-0.4ex]
 \includegraphics[width=10mm]{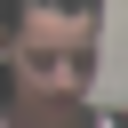} &  \includegraphics[width=10mm]{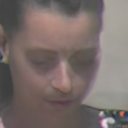}&  
\includegraphics[width=10mm]{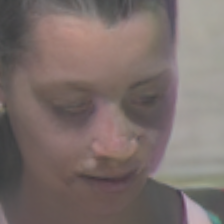}&\includegraphics[width=10mm]{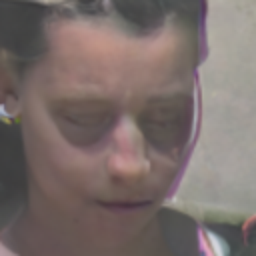}   &  \includegraphics[width=10mm]{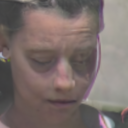} &  \includegraphics[width=10mm]{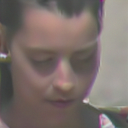} &\includegraphics[width=10mm]{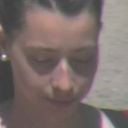} \\[-0.4ex]
\includegraphics[width=10mm]{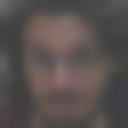} &  \includegraphics[width=10mm]{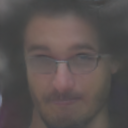}&  
\includegraphics[width=10mm]{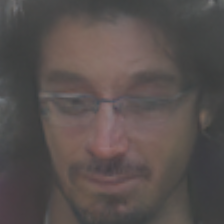}&\includegraphics[width=10mm]{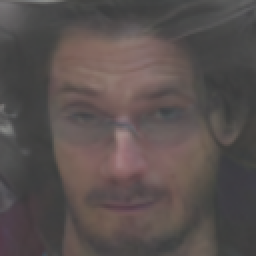}   &  \includegraphics[width=10mm]{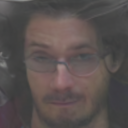} &  \includegraphics[width=10mm]{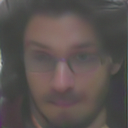}&\includegraphics[width=10mm]{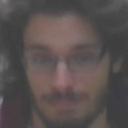}  \\[-0.4ex]
 \includegraphics[width=10mm]{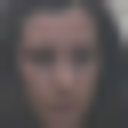} &  \includegraphics[width=10mm]{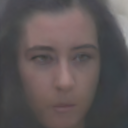}&  
\includegraphics[width=10mm]{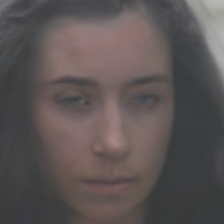}&\includegraphics[width=10mm]{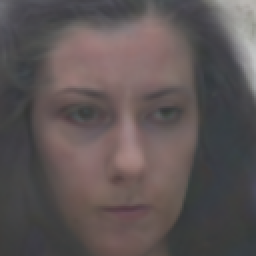}   &  \includegraphics[width=10mm]{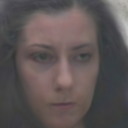} &  \includegraphics[width=10mm]{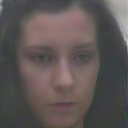} &\includegraphics[width=10mm]{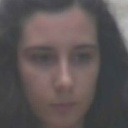} 
\end{tabular}
} 
\vspace{-4.5mm}
\caption{Comparison of low-resolution~(LR), super-resolution~(SR) results obtained by various methods, and ground gruth~(GT) images from the Quis-Campi dataset. \gls*{srname} outperforms baseline methods, preserving facial symmetry and natural appearance.}
\label{fig_quali}
\vspace{-1mm}
\end{figure*}

Lastly, the qualitative outcomes for on the Quis-Campi dataset are presented in Fig.~\ref{fig_quali}.
Our method, \gls*{srname}, is compared against other methods utilizing diffusion models to restore facial details and features.
While these methods are effective to some extent, they often introduce artifacts or noise onto the facial images, typical issues encountered in \gls*{sr} algorithms.
In contrast, \gls*{srname} stands out as the only approach that produces natural-looking images without noticeable artificiality.
It maintains facial naturalness, preserves symmetries, and successfully recovers details without introducing artifacts or distorting facial~features.

For instance, in row 3 of Fig.~\ref{fig_quali}, images generated by other algorithms exhibit distortions, particularly in the eye region, leading to a loss of naturalness and symmetry in the faces.
Conversely, in row 4 of Fig.~\ref{fig_quali}, our algorithm produces images with significantly reduced noise compared to the others.
Moving to row 6 of Fig.~\ref{fig_quali}, images generated by the SDE-SR and SRDG algorithms appear to ``age'' the subject, whereas our algorithm preserves the person's age while maintaining their identity without~distortion.

Due to the ill-posed nature of the SR problem, many SR algorithms suffer from bias issues and struggle to recover a person's identity accurately.
Our algorithm tackles this by fusing a reference LR image with a multi-feature vector, effectively mitigating identity-related problems and yielding superior quantitative and qualitative results.
However, additional tests and experiments are required to reduce bias and identity distortions before deploying this algorithm in real-world scenarios, especially in surveillance environments characterized by noisy and more challenging data, where errors in facial recognition can have adverse~consequences.

\subsection{Failure Cases}

\begin{figure}[!htb]
\vspace{-1.5mm}
\centering
\setlength{\tabcolsep}{1.2pt}
\resizebox{0.8\linewidth}{!}{ %
\begin{tabular}{ccc}
\tiny{SRDG~\cite{visapp24sr}}&\tiny{\gls*{srname} (Ours)} &\tiny{GT}\\[-0.3ex]
\includegraphics[width=15mm]{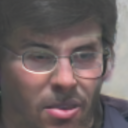} & 
\includegraphics[width=15mm]{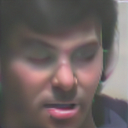} &\includegraphics[width=15mm]{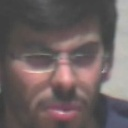} \\[-0.25ex]

\tiny{SDE-SR~\cite{dos2022face}}&\tiny{\gls*{srname} (Ours)} &\tiny{GT}\\[-0.3ex]
\includegraphics[width=15mm]{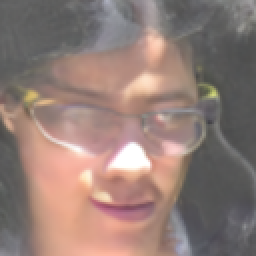} &
\includegraphics[width=15mm]{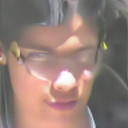} & 
\includegraphics[width=15mm]{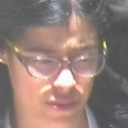} \\
\end{tabular}
}
\vspace{-1.5mm}
\caption{Failure cases: the first row presents results from SRDG, \gls*{srname}~(ours), and ground truth~(GT) images, while the second row presents results from SDE-SR, \gls*{srname}~(ours), and GT images.}
\label{fig_fail}
\end{figure}

Fig.~\ref{fig_fail} shows some failure cases of our algorithm compared to SRDG and SDE-SR.
In the first row, \gls*{srname} fails to recover the eyeglasses correctly, whereas SRDG successfully recovers this attribute.
Nevertheless, it is important to note that SRDG requires explicit information about whether the person is wearing eyeglasses.
This information is not always discernible from \gls*{lr} images in surveillance~scenarios.

In the second row of Fig.~\ref{fig_fail}, we observe a failure case of \gls*{srname} compared to SDE-SR.
The image in question shows significant pose variation and highly heterogeneous illumination.
\gls*{srname} produces smoother images with less noise than the other algorithms, causing the information about eyeglasses and the sun's reflection to spread across the periocular~region.

Upon closer examination of the cases where our algorithm fails in Rank-5, we observed that most images share characteristics similar to those described in the previous paragraphs.
Thus, \gls*{srname} provides better results for recognition accuracy but may be more sensitive to variations in pose and~lighting.

\section{Conclusions}
\label{sec:Conclusions}

In this work, we introduced \gls*{srname}, an algorithm that integrates multi-features and a reference LR image into diffusion models to generate SR images.
A key advantage of our algorithm is its independence from explicitly provided facial attributes; instead, it utilizes features extracted using a deep neural network.
This methodology enables our algorithm to preserve individuals' identities more effectively than other methods, resulting in high-quality SR images with enhanced face symmetry, reduced noise and minimized distortions in face attributes.
Our approach was validated on the CelebA and Quis-Campi datasets, where we achieved state-of-the-art results for recognition metrics. 
Hence, it demonstrates the potential to be applied in real-world surveillance scenarios.

\section*{\uppercase{Acknowledgments}}

\iffinal
    This study was financed in part by the \textit{Coordenação de Aperfeiçoamento de Pessoal de Nível Superior - Brasil~(CAPES)} - Finance Code 001, and in part by the \textit{Conselho Nacional de Desenvolvimento Científico e Tecnológico~(CNPq)} (\#~315409/2023-1).
    We thank the support of NVIDIA Corporation with the donation of the Quadro RTX $8000$ GPU used for this research.
\else
    The acknowledgments are hidden for review.
\fi

\balance

\bibliographystyle{IEEEtran}
\bibliography{bibtex}

\end{document}